\documentclass{iopjournal}
\usepackage{hyperref}
\usepackage{url}
\usepackage{graphicx}
\usepackage{amsmath}
\usepackage{algorithm}
\usepackage{algpseudocode}
\usepackage{graphicx}
\usepackage{multirow}
\usepackage{booktabs}
\usepackage{pgfplots}
\usepackage{natbib}
\pgfplotsset{compat=1.18}
\usetikzlibrary{pgfplots.groupplots}
\usepackage{tikz}
\usepackage{hyperref}
\usepackage{fontawesome5}

\makeatletter
\renewcommand{\articletype}[1]{{\scriptsize \sf{\bfseries \MakeUppercase{#1}}}}
\makeatother

\begin{document}

\articletype{Paper} 

\title{Principal Prototype Analysis on Manifold for Interpretable Reinforcement Learning}

\author{Bodla Krishna Vamshi$^1$ and Haizhao Yang$^2$ }

\affil{$^1$Department of Computer science, University of Maryland College park, USA}

\affil{$^2$Department of Mathematics, University of Maryland College park, USA}




\begin{abstract}
Recent years have witnessed the widespread adoption of reinforcement learning (RL), from solving real-time games to fine-tuning large language models using human preference data significantly improving alignment with user expectations. However, as model complexity grows exponentially, the interpretability of these systems becomes increasingly challenging. While numerous explainability methods have been developed for computer vision and natural language processing to elucidate both local and global reasoning patterns, their application to RL remains limited. Direct extensions of these methods often struggle to maintain the delicate balance between interpretability and performance within RL settings. Prototype-Wrapper Networks (PW-Nets) have recently shown promise in bridging this gap by enhancing explainability in RL domains without sacrificing the efficiency of the original black-box models. However, these methods typically require manually defined reference prototypes, which often necessitate expert domain knowledge. In this work, we propose a method that removes this dependency by automatically selecting optimal prototypes from the available data. Preliminary experiments on standard Gym environments demonstrate that our approach matches the performance of existing PW-Nets, while remaining competitive with the original black-box models.
\end{abstract}

\section{\textbf{Introduction}}
Deep reinforcement learning (RL) models have achieved state-of-the-art performance in domains such as Go \citet{44806}, Chess \citet{silver2017masteringchessshogiselfplay}, inverse scattering \citet{jiang2022reinforcedinversescattering}, and self-driving cars \citet{kiran2021deepreinforcementlearningautonomous}. More recently, RL has been successfully applied to align large language models with human preferences, receiving considerable attention as a powerful post-training strategy using extensive human feedback data  \citet{ouyang2022traininglanguagemodelsfollow, rafailov2024directpreferenceoptimizationlanguage}. However, despite these advances, the deployment of RL agents in sensitive domains remains limited due to the opaque nature of their decision-making processes. Extracting the rationale behind an agent’s actions in a human-interpretable format remains a significant challenge, yet doing so is crucial for understanding failure modes and ensuring trust in these systems. To address this challenge, prototype-based networks have emerged as a promising approach for enhancing the interpretability of deep learning models. ProtoPNet \citet{chen2019lookslikethatdeep}, initially proposed for image classification tasks, introduced pre-hoc interpretability by associating predictions with learned prototype representations. 

This idea was later extended to deep RL with Prototype-Wrapper Networks (PW-Nets) \citet{kenny2023towards}, which provide post-hoc interpretability while preserving the performance of the underlying black-box agent. By incorporating exemplar-based reasoning, PW-Nets allow users to inspect and understand the agent’s actions through user-defined reference examples, without degrading task performance. Despite these recent advantages, there is a remaining challenge to automatically and efficiently discover data-adaptive reference examples for interpreting RL behaviors, since manually curated prototypes present several limitations: Human-selected prototypes are costly to acquire, difficult to scale, and often lack consistency across environments, reducing the reproducibility and generalization of explanations. To overcome the above limitations, we propose our principal prototype analysis on manifold: an automated prototype sampling method that eliminates the need for manual intervention and selects prototypes adaptive to RL tasks on the data manifold. To the best of our knowledge, this is among the first works to explore automated prototype discovery in reinforcement learning settings while retaining the performance of the black-box agent. Our approach leverages a combination of metric and manifold learning objectives to select prototypes directly from the encoded state space that reflects a low-dimensional geometric representation of the RL task, providing a more scalable and principled mechanism for prototype discovery. The code is publicly available at \href{https://github.com/strangeman09/principal_prototype_analysis}{\faGithub\ Code}

\begin{itemize}
    \item \textbf{Automated and Decoupled Prototype Discovery:} Our method proposes a novel two-stage architecture that decouples prototype discovery from policy optimization. In the first stage, it automatically selects prototypes from the agent’s trajectory data using a lightweight neural network trained with combined manifold and metric learning objectives, removing the need for human-curated examples. In the second stage, these prototypes are fixed and integrated into the PW-Net for interpretable action prediction, preserving black-box performance.

    \item \textbf{Geometry-Aware and Faithful Prototypes via Real Instances:} Instead of learning abstract embeddings, our method grounds each learned proxy vector in real training samples by mapping them to their nearest encoded instance. This ensures prototypes are both geometry-aware by leveraging piecewise-linear manifold approximations and semantically faithful, enabling more intuitive and interpretable behavior analysis of RL agents.
\end{itemize}

\section{\textbf{Related Works}}

\label{sec:Related Works}
Interpretability in neural network architectures, particularly in computer vision (CV) and natural language processing (NLP), has advanced substantially, encompassing both pre-hoc and post-hoc strategies. In CV, post-hoc methods such as Grad-CAM \citet{Selvaraju_2019}, RISE \citet{petsiuk2018riserandomizedinputsampling}, and occlusion-based techniques like Meaningful Perturbations \citet{Fong_2017} have enabled visual explanations by highlighting image regions most influential to predictions. However, these methods provide explanations only after decisions are made, offering limited insight into the decision-making process itself. In NLP, pre-hoc approaches include interpretable rule-based decision sets \citet{10.1145/2939672.2939874} and, more recently, Proto-LM \citet{xie2023protolmprototypicalnetworkbasedframework}, which embeds prototypical reasoning directly into large language models. Post-hoc methods such as LIME \citet{ribeiro2016whyitrustyou} and Integrated Gradients \citet{sundararajan2017axiomaticattributiondeepnetworks} are widely used to approximate local model behavior and attribute predictions to input features. Other efforts have challenged conventional practices; for instance, \citet{jain2019attentionexplanation} questioned the reliability of attention weights as explanations, while \citet{arras2016explainingpredictionsnonlinearclassifiers} applied Layer-wise Relevance Propagation to trace decision origins in text classifiers.

Although several interpretability techniques have been proposed for reinforcement learning (RL) models \citet{Vouros_2022,milani2022surveyexplainablereinforcementlearning}, most prior work relies on interpretable surrogate models, such as decision trees, that imitate agent behavior in symbolic domains. These approaches, however, do not scale to complex environments with high-dimensional observations such as high-dimensional pixel-based observations. In deep RL settings, most interpretability research has focused on post-hoc methods utilizing attention mechanisms \citet{zambaldi2018deep,mott2019interpretablereinforcementlearningusing} or tree-based surrogates \citet{liu2018interpretabledeepreinforcementlearning}, but these often fall short in revealing the underlying reasoning or intent of the agent \citet{rudin2021interpretablemachinelearningfundamental}. Some approaches attempt to distill recurrent neural network (RNN) policies into finite-state machines \citet{danesh2021reunderstandingfinitestaterepresentationsrecurrent,koul2018learningfinitestaterepresentations}, but such methods can yield opaque explanations and are constrained to specific architectures.

Our work builds on prototype-based neural networks, which are inherently interpretable by design. These models associate test instances with prototypical examples during the forward pass, enabling intuitive, exemplar-based reasoning. A foundational example of this approach was presented by \citet{li2017deeplearningcasebasedreasoning}, who introduced a pre-hoc method that learns prototypes in latent space and classifies inputs based on their L2 distance to these prototypes. This method also required a decoder to visualize prototype representations. A notable extension was ProtoPNet \citet{chen2019lookslikethatdeep}, which associated prototypes with image parts rather than entire images, enhancing fine-grained interpretability.

The prototype network paradigm has been substantially extended since ProtoPNet, with several lines of work addressing its core limitations. ProtoTree \citet{nauta2021neuralprototypetreesinterpretable} embedded prototype reasoning within a hierarchical decision tree structure, enabling global explanations through a sequence of prototype comparisons rather than a single similarity score. PIPNet \citet{Nauta_2023_CVPR} further improved visual coherence by enforcing that prototypes correspond to contiguous, human-interpretable image patches rather than distributed feature vectors. In the few-shot learning literature, Prototypical Networks \citet{snell2017prototypicalnetworksfewshotlearning} demonstrated that class-level prototype representations computed as the mean of support set embeddings provide a powerful inductive bias for generalization, establishing the theoretical basis for prototype-based reasoning in low-data regimes. Building on this, PAL \citet{arik2019protoattendattentionbasedprototypicallearning} introduced prototype-based attentive learning to dynamically weight prototype contributions, improving robustness to intra-class variation. A common limitation across these methods is that prototypes are either learned end-to-end alongside the classification objective, risking a performance-interpretability tradeoff, or fixed as simple class statistics such as means or medoids, which fail to capture the intrinsic geometry of the learned representation space. Our method addresses this gap by decoupling prototype discovery from task optimization and grounding prototype selection in the manifold structure of the encoded state space, ensuring that prototypes are both geometrically faithful and discriminative without degrading task performance.

A related but distinct line of interpretability research focuses on concept-based explanations, which seek to explain model behavior in terms of high-level semantic concepts rather than individual training instances. Testing with Concept Activation Vectors (TCAV) \citet{kim2018interpretabilityfeatureattributionquantitative} quantifies the sensitivity of model predictions to user-defined concept directions in activation space, enabling global explanations without modifying the model architecture. Concept Bottleneck Models (CBMs) \citet{koh2020conceptbottleneckmodels} extend this idea by explicitly conditioning predictions on a learned concept layer, allowing human intervention at inference time. More recent work such as ConceptSHAP \citet{yeh2022completenessawareconceptbasedexplanationsdeep} has sought to automate concept discovery by identifying a complete and minimal set of concepts that explain model behavior without requiring predefined concept annotations. While these approaches provide semantically rich explanations, they share two limitations that constrain their applicability to RL settings: they typically require either predefined concept vocabularies or auxiliary concept-annotated datasets, and they do not ground explanations in concrete training instances, making it difficult for non-expert users to interpret agent behavior through direct analogy. Our prototype-based approach complements concept-level methods by providing instance-level, exemplar-based explanations that are immediately interpretable without domain-specific concept supervision, while the geometry-aware selection mechanism ensures that chosen exemplars faithfully represent the decision-relevant structure of the agent's learned state representation.

\section{\textbf{Methodology}}

\subsection{\textbf{Motivation}}
\label{sec:motivation}
Prototype-based methods offer an interpretable way to associate each class with representative examples; here the representative examples are termed as prototypes. A straightforward baseline to define prototypes is using simple statistics such as the class mean or medoids in the embedding space. However, such naive approaches fail to capture the intrinsic geometry of encoded representations: they are biased by outliers, insensitive to multi-modal distributions within classes, and often yield prototypes that are statistically central but semantically uninformative. To construct meaningful prototypes, it is essential to account for the geometry of the data distribution itself.

According to the manifold hypothesis ~\cite{article}, high-dimensional representations typically reside on lower-dimensional manifolds. Leveraging this property enables geometry-aware prototype sampling. Classical manifold learning techniques, however, come with limitations methods like t-SNE \citet{JMLR:v9:vandermaaten08a}, UMAP \citet{mcinnes2020umapuniformmanifoldapproximation}, and LLE \citet{doi:10.1126/science.290.5500.2323} emphasize neighborhood preservation but often distort local dependencies or fail to provide consistent global structure. To address this, we adopt a piecewise-linear manifold learning approach in which nonlinear manifolds are decomposed into locally linear regions. This design ensures that prototypes are drawn from regions that reflect local geometry, avoiding the pitfalls of global averages or distorted embeddings.

While manifold learning preserves geometric structure, prototypes must also be discriminative across classes. Geometry alone does not guarantee that prototypes tightly capture intra-class consistency or maximize inter-class separation. To achieve this, we incorporate a metric learning objective. Methods such as triplet or contrastive loss require predefined prototypes and extensive sample mining, which is inefficient and often unstable. Instead, we employ Proxy-Anchor loss, which introduces learnable class-level proxy vectors that directly enforce compact clustering within a class and clear separation between classes. After training, each proxy is mapped to its nearest training instance, yielding prototypes that are simultaneously geometry-aware and discriminative.

In \citet{chen2019lookslikethatdeep}, the notion of learnable prototypes was introduced for image classification, where prototype learning was jointly optimized alongside the classification objective. While this approach proved effective for supervised image tasks, its adaptation to reinforcement learning in \citet{kenny2023towards} (PW-Net*) resulted in noticeably weaker performance compared to black-box RL models. To overcome this limitation, we propose to decouple these objectives into two sequential stages. In the first stage, we focus on sampling prototypes that serve as robust and representative anchors for each class. In the second stage, these prototypes are fixed and used within PW-Net, which is then trained exclusively on the RL objective.

\subsection{\textbf{Dataset}}
\label{subsec:dataset}

Our method begins with the assumption that we have access to a pre-trained policy $\pi_{\text{bb}}$ operating within a Markov Decision Process (MDP) \citet{sutton2015rl}. Since all policies used in our experiments are implemented as neural network architectures, we assume that each policy concludes with a final linear layer. Under this setting, the policy $\pi_{\text{bb}}$ can be decomposed into two components: an encoder $f_{\text{enc}}$, which maps the input state $s$ to a latent representation $z$, and a final linear layer defined by weights $W$ and bias $b$. The resulting policy function can be expressed as:
\[
    \pi_{\text{bb}}(s) = W f_{\text{enc}}(s) + b,
\]
Where $z = f_{\text{enc}}(s)$ represents the encoded state.To construct the dataset used for training our prototype selection mechanism, we execute the pre-trained agent in its original environment for $n$ time steps. During this rollout, we collect encoded state--action pairs, resulting in a dataset $D$:
\[
    D \leftarrow \{(z_i, \pi_{\text{bb}}(s_i)) \}_{i=1}^{n}.
\]

\subsection{\textbf{Training overview}}

\begin{figure}[htbp!]
    \centering
    \includegraphics[width=\linewidth]{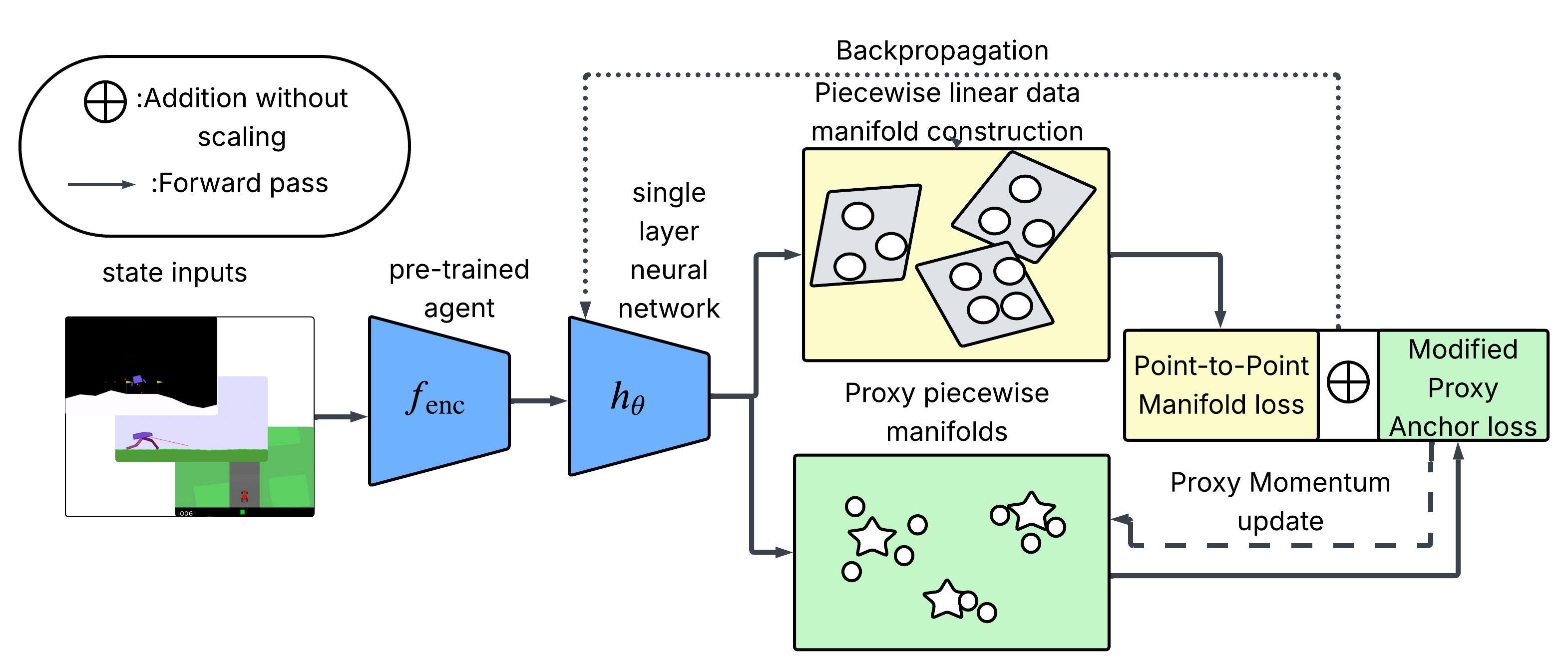}
    \caption{Overview of the proposed method
    }
    \label{fig:method_overview}
\end{figure}

As mentioned in ~\ref{sec:motivation}, our method consists of two stages. In the first stage of our method, we initialize a simple neural network $h_\theta$ and train it on Dataset $D$ to jointly optimize manifold learning Eq.:~\ref{eq:manifold_loss} and metric learning objectives Eq.:~\ref{eq:pca}. The neural network $h_\theta$ learns to map the high-dimensional encoded representations into lower dimensions. Before the training process, we initialize the proxies $\theta_q$ and $\theta_m$; here both the proxies are unique for each class and initiated randomly with $\theta_q$ = $\theta_m$.
The proxy vector $\theta_q$ is learned using the metric learning objective Eq.:~\ref{eq:pca} and updated via back-propagation. The proxy vector $\theta_m$ is updated via the Momentum update \citet{he2020momentumcontrastunsupervisedvisual} where $\gamma$ is the momentum constant.
\begin{equation}
 \label{eq:momentum_update}
\theta_m \leftarrow \gamma\theta_m + (1 - \gamma)\theta_q
\end{equation}

Before training our model $h_\theta$, we reformat the dataset $D$ to consist of pairs of encoded state representations and their corresponding discretized actions (Section~\ref{subsec:discritization}). This discretization allows the use of a metric learning objective Eq.:~\ref{eq:pca} that clusters encoded states with similar actions and separates those with dissimilar ones, and also enables learning discriminative prototypes.

During training, for every mini-batch $B$ we build linear piecewise manifolds as outlined in Section ~\ref{sec:manifold_construction}. For every point in $B$, we then compute the manifold-based similarity following the procedure in Section ~\ref{para:manifold_sim}. This similarity measure is used to compute the manifold point-to-point loss $\mathcal{L}_{\text{manifold}}$. At the same time, we compute the Proxy Anchor loss $\mathcal{L}_{\text{PA}}$ using randomly initialized class proxies $\theta_q$ and latent representations $z$ in batch $B$. The final loss is computed as $\mathcal{L}_{\text{total}} = \mathcal{L}_{\text{PA}} + \mathcal{L}_{\text{manifold}}$. 

The manifold point-to-point loss is designed to reduce the distance between points lying on the same manifold, thus preserving local geometric structure while increasing the distance between points on different manifolds. In contrast, the Proxy Anchor loss encourages samples from the same class to cluster closer together while pushing samples from different classes further apart; this encourages the discriminative learning of prototypes. For every epoch, the network $h_\theta$ is updated through backpropagation, and the proxy vectors are updated according to the procedure described in Eq.: ~\ref{eq:momentum_update}. Once the training is completed, we use the learned proxy vectors $\theta_m$  to select the nearest training data sample as prototypes for each class to be used in the stage of two of training PW-net; here, for every class, there is only $\theta_m$ being initialized, i.e., we will be only getting one prototype per class.

\subsection{\textbf{Manifold construction}}
\label{sec:manifold_construction}

Based on the Manifold hypothesis, we assume that the encoded state representations produced by the policy $\pi_{\text{bb}}$, though inherently complex and non-linear, can be locally approximated into smaller chunks of linear regions. Our approach leverages this structural assumption to automatically identify representative prototypes that capture the essential characteristics of each action class.


To efficiently approximate the structure of the data manifold, we adopt a piecewise linear manifold learning method, which constructs localized $m$-dimensional linear submanifolds around selected anchor points. Given a batch $B$ containing $N$ data points, we randomly select $n$ of them to serve as anchors. For each anchor point $h_\theta(z_i)$, we initially collect its $m{-}1$ nearest neighbors in the encoded representation space based on Euclidean distance to form the neighborhood set $X_i$.

The manifold expansion process proceeds iteratively by attempting to add the $m$-th nearest neighbor to $X_i$. After each addition, we recompute the best-fit $m$-dimensional submanifold using PCA and assess whether all points in $X_i$ can be reconstructed with a quality above a threshold $T\%$. If the reconstruction quality remains acceptable, the new point is retained in $X_i$; otherwise, it is excluded. This evaluation is repeated for subsequent neighbors $N(h_\theta(x_i))_j$ for $j \in \{m_l+1, \dots, k\}$, gradually constructing a local linear approximation of the manifold.

The final set $X_i$ comprises all points in the anchor's neighborhood that lie well within an $m$-dimensional linear submanifold. A basis for this submanifold is computed by applying PCA to $X_i$ and extracting the top $m$ eigenvectors. We choose PCA for this task as it is computationally efficient and well-suited for capturing linear approximations of non-linear data, in alignment with our assumption of locally linear structure within the high-dimensional state space.

\subsection{\textbf{Loss Functions}}
\paragraph{Proxy Anchor Loss:}
We use a modified version of proxy anchor loss with Euclidean distance instead of cosine similarity:
\begin{align}
\label{eq:pca}
\mathcal{L}_{\text{PA}} = 
&\frac{1}{|\Theta_+|} \sum_{\theta_q \in \Theta_+} 
\log\left(1 + \sum_{z \in \mathcal{Z}_{\theta_q}^+} 
\exp\left(-\alpha \cdot \left(\|h_\theta(z) - \theta_q\|_2 - \epsilon\right)\right)\right) \\
+\, 
&\frac{1}{|\Theta|} \sum_{\theta_q \in \Theta} 
\log\left(1 + \sum_{z \in \mathcal{Z}_{\theta_q}^-} 
\exp\left(\alpha \cdot \left(\|h_\theta(z) - \theta_q\|_2 - \epsilon\right)\right)\right)
\end{align}

Here, $\Theta$ denotes the set of all proxies, where each proxy 
$\theta_q \in \Theta$ serves as a representative vector for a class. 
The subset $\Theta_+ \subseteq \Theta$ includes only those proxies 
that have at least one positive embedding in the current batch $B$. 
For a given proxy $\theta_q$, the latent representations $\mathcal{Z}$ in $B$ (where $z \in \mathcal{Z}$)
are partitioned into two sets: $\mathcal{Z}_{\theta_q}^+$, the positive embeddings 
belonging to the same class as $\theta_q$, and 
$\mathcal{Z}_{\theta_q}^- = \mathcal{Z} \setminus \mathcal{Z}_{\theta_q}^+$, 
the negative embeddings. The scaling factor $\alpha$ controls the sharpness of optimization by amplifying hard examples when large (focusing gradients on difficult pairs) or smoothing training when small (spreading weight across all pairs). The margin 
$\epsilon$ enforces a buffer zone between positives and negatives by requiring positives to be closer to their proxies and negatives to be sufficiently farther away. 

\paragraph{Manifold Point-to-Point Loss:}
\label{para:manifold_sim}
This loss helps in estimating the point to point similarities preserving the geometric structure:

\begin{equation}
\label{eq:manifold_loss}
\small
\mathcal{L}_{\text{manifold}} = \sum_{i,j} \left(\delta \cdot (1 - s(z_i, z_j)) - \|h_\theta(z_i) - h_\theta(z_j)\|_2\right)^2
\end{equation}

\noindent where \( s(z_i, z_j) \) is the manifold similarity computed as:
\begin{equation*}
\small
s(z_i, z_j) = \frac{s'(z_i, z_j) + s'(z_j, z_i)}{2}
\end{equation*}

\noindent with \( s'(z_i, z_j) = \alpha(z_i, z_j) \cdot \beta(z_i, z_j) \), where:
\begin{equation*}
\small
\begin{aligned}
\alpha(z_i, z_j) &= \frac{1}{\left(1 + o(z_i, z_j)^2\right)^{N_\alpha}} \\
\beta(z_i, z_j)  &= \frac{1}{\left(1 + p(z_i, z_j)\right)^{N_\beta}}
\end{aligned}
\end{equation*}

$\delta$ is the scaling factor, it determines the maximum separation between dissimilar points. The loss encourages Euclidean distances in the embedding space to match manifold-based dissimilarities $1-s(z_i, z_j)$, ensuring that the learned metric space respects the underlying manifold structure.
 $o(z_i, z_j)$ is the orthogonal distance from point $z_i$ to the manifold of point $z_j$, and $p(z_i, z_j)$ is the projected distance between point $z_j$ and the projection of $z_i$ on the manifold. The parameters $N_\alpha$  and $N_\beta$ control how rapidly similarity decays with distance, with $N_\alpha > N_\beta$ ensuring that similarity decreases more rapidly for points lying off the manifold than for points on the same manifold.

 \paragraph{Distance Calculation.}
For each point pair $(z_i, z_j)$, the distances $o(z_i, z_j)$ and $p(z_i, z_j)$ are calculated using the manifold basis vectors $P_j$ associated with point $z_j$. The projection of $z_i$ onto $P_j$ is computed as $\text{proj}_{P_j}(z_i) = z_j + \sum_{k} \langle z_i - z_j, v_k \rangle v_k$, where $v_k$ are the basis vectors of $P_j$. The orthogonal distance is then $o(z_i, z_j) = \|z_i - \text{proj}_{P_j}(z_i)\|_2$, and the projected distance is $p(z_i, z_j) = \|\text{proj}_{P_j}(z_i) - z_j\|_2$. This process is repeated for all point pairs, capturing the full geometric structure of the data manifold.

The total loss is the sum of these two components, allowing the model to simultaneously learn a metric space that respects action classes while preserving the geometric structure of the data.

\subsection{\textbf{Performance review}}

The action output $a'$ from the Prototype-Wrapper Network (PW-Net) can in some cases generalize better than the original black-box model’s action $a$ \citet{snell2017prototypicalnetworksfewshotlearning,li2021adaptiveprototypelearningallocation}, due to improved alignment with class-representative prototypes—even without further interaction with the environment. This generalization is critically influenced by the quality and representativeness of the selected prototypes.
The black-box policy $\pi_{bb}$ computes the action as:
\begin{equation*}
    a = W f_{\text{enc}}(s) + b
\end{equation*}
where $z$ is the latent state representation obtained from the encoder. PW-Net enforces structured reasoning through prototypes and computes similarity scores as:
\begin{equation*}
    a'_i = \sum_{j=1}^{N_i} W'_{i,j} \text{sim}(z_{i,j}, p_{i,j})
\end{equation*}
The similarity function is defined as:
\begin{equation*}
    \text{sim}(z_{i,j}, p_{i,j}) = \log \left( \frac{(z_{i,j} - p_{i,j})^2 + 1}{(z_{i,j} - p_{i,j})^2 + \epsilon} \right).
\end{equation*}
This ensures actions are chosen based on structured prototype distances rather than raw neural activations. The model uses prototype-based regularization, which provides better generalization by using the  learned policy $\pi_{\text{bb}}$ as additional input signal. For simplicity assume a deep RL domain with only two actions possible, the action can be computed as $a'$
\begin{equation*}
   a'_1 = W'_{1,1} \log \left( \frac{d_{1,1}^2 + 1}{d_{1,1}^2 + \epsilon} \right) + W'_{1,2} \log \left( \frac{d_{1,2}^2 + 1}{d_{1,2}^2 + \epsilon} \right)
\end{equation*}

\begin{equation*}
   d_{i,j} = z_{i,j} - p_{i,j}.
\end{equation*}
Where $W'$ is the manually defined weight matrix for each action, the output $a'$ is heavily dependent on the similarity score between the $z_{i,j}$ and $p_{i,j}$, this metric enables PW-Net to avoid completely mimicking the policy $\pi_{bb}$ and instead use it as an additional input signal along with the choice of prototype to better align responses with human choices. 

\subsection{\textbf{Model Architecture}}
This section includes details about the black-box models, user study, and the model architecture ($h_\theta$) used in our method. We used a single-layer network with intermediate normalizations. The prototype size is set to 50 for all the environments. The motivation for using a simpler model is to avoid losing information in the encoded vectors during manifold construction.

\begin{table}[h!]
\caption{Model Architecture}
\label{tab:transformation_architecture}
\centering
\begin{tabular}{@{}ll@{}}
\toprule
\textbf{Layer} & \textbf{Layer Parameters} \\
\midrule
Linear & (latent size $z$, prototype size) \\
InstanceNorm1d & prototype size  \\
ReLU & - \\
\bottomrule
\end{tabular}
\end{table}

\section{\textbf{Experiments}}

\subsection{\textbf{Action Discretization}}
\label{subsec:discritization}

Our prototype discovery framework requires discrete class labels to enable metric learning and clustering of encoded state representations. While such labels are naturally available in discrete-action environments, continuous control settings lack an explicit class structure. To address this, we discretize the continuous action space by mapping each action vector to a single dominant action class. Specifically, we transform the action values using a sigmoid function and assign each encoded state to the class corresponding to the maximum activation. This process identifies the most influential control dimension at each state, thereby preserving the primary decision signal of the policy.

We use absolute values of the action components prior to transformation to emphasize action magnitude rather than direction. This design choice aligns with our objective of capturing dominant behavioral modes for interpretability, rather than modeling fine-grained directional variations in control. For instance, in the Car Racing environment, the original action output is represented as a tuple \texttt{[(acc, brake), left, right]}. We first restructure this into a unified vector format: \texttt{[acc, brake, left, right]}. The encoded state representation is then assigned a discrete label based on the index of the maximum value obtained after applying the sigmoid function to this transformed vector.

This discretization procedure is consistently applied across all continuous action environments, including the Bipedal Walker and Humanoid Standup environments, enabling compatibility with our prototype selection and metric learning pipeline. Importantly, our goal is not to exactly reconstruct the continuous action space, but to enable interpretable grouping of states based on dominant action tendencies. This discretization provides a consistent and scalable mechanism for prototype discovery across both discrete and continuous domains, while maintaining alignment with the agent’s decision structure. Empirically, we observe that this transformation does not degrade policy performance, as demonstrated in Section ~\ref{subsec:results}.

\subsection{\textbf{Numerical Results}}
\label{subsec:results}
\begin{table}[h]
\centering
\small
\setlength{\tabcolsep}{7pt} 
\begin{tabular}{l|c|c|c}
\toprule
\textbf{Method} & \textbf{Car Racing} & \textbf{BipedalWalker} & \textbf{HumanoidStandup} \\
& \textbf{(Reward)} & \textbf{(Reward)} & \textbf{(Reward)} \\
\midrule
Our method & \textbf{220.91 $\pm$ 0.85} & 312.10 $\pm$ 0.17 & \textbf{75112.60 $\pm$ 840.25} \\
PW-Net & 220.72 $\pm$ 0.34 & 308.27 $\pm$ 3.41 & 74980.37 $\pm$ 816.84 \\
VIPER & N/A & -92.36 $\pm$ 10.09 & - \\
PW-Net* & -10.23 $\pm$ 2.20 & 197.85 $\pm$ 52.19 & - \\
k-means & -1.56 $\pm$ 0.81 & -105.68 $\pm$ 0.54 & - \\
Black-Box  & 219.56 $\pm$ 0.85 & \textbf{312.32 $\pm$ 0.21} & 74930.50 $\pm$ 837.61 \\
\bottomrule
\end{tabular}
\caption{Reward comparison on Car Racing, Bipedal Walker, and Humanoid Standup tasks}
\label{tab:results1}
\end{table}

\begin{table}[h]
\centering
\small
\setlength{\tabcolsep}{7pt}
\begin{tabular}{l|c|c|c}
\toprule
\textbf{Method} & \textbf{Pong (Reward)} & \textbf{Lunar Lander (Reward)} & \textbf{Acrobot (Reward)} \\
\midrule
Our method & \textbf{14.96 $\pm$ 0.45} & \textbf{218.01 $\pm$ 1.47} & \textbf{-83.12 $\pm$ 2.39} \\
PW-Net & 10.72 $\pm$ 0.26 & 216.38 $\pm$ 1.69 & -84.67 $\pm$ 2.42 \\
VIPER & N/A & -423.22 $\pm$ 53.91 & - \\
PW-Net* & 8.64 $\pm$ 1.27 & 131.01 $\pm$ 89.52 & - \\
k-means & -19.47 $\pm$ 2.49 & -423.41 $\pm$ 97.85 & - \\
Black-Box & 12.07 $\pm$ 0.39 & 214.75 $\pm$ 1.08 & -85.54 $\pm$ 3.37 \\
\bottomrule
\end{tabular}
\caption{Reward comparison on Pong, Lunar Lander, and Acrobot environments.}
\label{tab:results2}
\end{table}

Tables ~\ref{tab:results1} and ~\ref{tab:results2} present the performance comparison across six reinforcement learning environments. All methods are evaluated under a controlled setting using identical encoder architectures (black-box), training procedures, hyperparameters (including epochs), and random seeds. The only difference between methods lies in the prototype selection strategy. Our method consistently matches or outperforms PW-Net across all environments, while achieving performance comparable to the original black-box policies. This demonstrates that automated prototype selection can improve interpretability without sacrificing policy performance. We do not report results for VIPER, PW-Net*, and k-means in certain high-dimensional environments (e.g. HumanoidStandup and Acrobot), as these methods fail to scale to such settings and do not produce meaningful policies in preliminary experiments.

The PW-Net \citet{kenny2023towards} relied on human-curated prototypes in visually interpretable environments such as Car Racing. However, this approach becomes infeasible in complex domains with high-dimensional, non-visual state spaces and large continuous action sets. For instance, the Humanoid Standup environment (Section ~\ref{sec:black_box_models}) features a high-dimensional vector input and 17 continuous control actions across joints and rotors, making manual prototype selection impractical without domain-specific tools or expertise. Our automated prototype selection method overcomes this limitation by leveraging geometric and class-level structure in the latent space. Notably, in the Humanoid Standup task, our approach achieves a mean reward of 75,112.60 (SE = 840.25), closely matching the original black-box model's performance of 74,930.50 (SE = 837.61). This result demonstrates that our method retains performance even in settings where manual prototype curation is infeasible. For the new environments (Humanoid Standup and Acrobot) without manually curated prototypes, we use class-mean prototypes as a reproducible heuristic baseline. This does not reflect the original PW-Net design but provides a consistent comparison in settings where manual selection is infeasible. This follows the approach used by the PW-Net authors (Section ~\ref{sec:method_comparison}) that they have used for training on the BipedalWalker and Lunar Lander environments.

To analyze the effect of varying hyperparameters, we have performed an ablation study (Section ~\ref{sec:ablation_study}) on the Bi-pedal and Atari Pong environments. To ensure a fair comparison, PW-Net and our method share the same encoder architecture, training procedure, hyperparameters (including learning rate, number of epochs), and random seeds. The only difference between the two methods lies in prototype selection: PW-Net relies on manually or heuristically selected prototypes, whereas our method automatically discovers prototypes through the proposed geometry-aware approach. This controlled setup isolates the effect of prototype selection on performance and interpretability.

While the original PW-Net work evaluates on four environments with manually selected prototypes, we extend the evaluation to six environments. For environments without predefined prototypes, we employ class-mean representations as a standardized baseline. This choice ensures reproducibility and reflects realistic settings where expert-defined prototypes are unavailable. Importantly, this highlights a key limitation of PW-Net—its reliance on manual prototype selection—which our method addresses through automated, geometry-aware prototype discovery.

\subsection{\textbf{Black-Box Models}}
\label{sec:black_box_models}
For the CarRacing environment, we used a CNN model trained using PPO \citet{jinayjain2025deep_racing}. This pre-trained model was evaluated under both IID and OOD settings during the user study. For Atari Pong, we used a simple CNN trained with the Double Dueling DQN method \citet{bhctsntrk_OpenAIPongDQN_2025}. The model used for BipedalWalker was trained using TD3 \citet{nikhilbarhate99_td3_pytorch_bipedalwalker_v2_2025}, and the LunarLander model was trained using the Actor-Critic method \citet{nikhilbarhate99_actorcritic_pytorch_2025}. These networks are relatively simple, reflecting the symbolic nature of their respective environments. For the HumanoidStandup and CartPole environments, we used models from Stable-Baselines3 \citet{stable-baselines3}, trained using PPO with an MLP policy. The diversity of environments, models, and algorithms demonstrates the robustness of our approach.

\subsection{\textbf{Training parameters}}
For the first phase of training prototype discovery we train our network for 200 epochs on the training dataset (Section~\ref{subsec:dataset}) using two separate Adam optimizers: one for the network parameters and one for the proxy parameters. Both optimizers use a learning rate of \texttt{1e-3}, accompanied by a learning rate scheduler with decay rate $\eta_t = 0.97$. The dimensionality of the encoded vector $z$ varies depending on the environment and the encoder model, but generally falls near the order of  100. We use a mini-batch size of 128 samples and set the reconstruction threshold $T$ to 90\%. The scale parameter $\delta$ is set to 2 (the maximum distance between two points on a unit sphere), and the submanifold dimension $m$ is fixed at 3.

For the second phase training and evaluating the sampled prototypes within the PW-Net framework we use the Adam optimizer with a learning rate of \texttt{1e-2}, again paired with a scheduler using $\eta_t = 0.97$. Training and evaluation are conducted over 5 independent iterations. In each iteration, the PW-Net model is trained for 10 epochs and evaluated over 30 simulation runs to compute the mean and standard deviation of the resulting rewards.

All experiments were conducted on an NVIDIA RTX A6000 GPU. In the first stage of our method, we train a lightweight neural network $h_\theta$ to sample prototypes, which requires approximately 640 MB of GPU memory and about 7 hours of training time without parallelization. With parallelized estimation of manifold-based similarities, the training time is reduced to roughly 2 hours, with a peak GPU memory usage of about 4700 MB across all environments. 

\subsection{\textbf{User Study}}
\begin{figure}[h]
    \centering
    \label{fig:user_study}
    \includegraphics[width=0.85\linewidth]{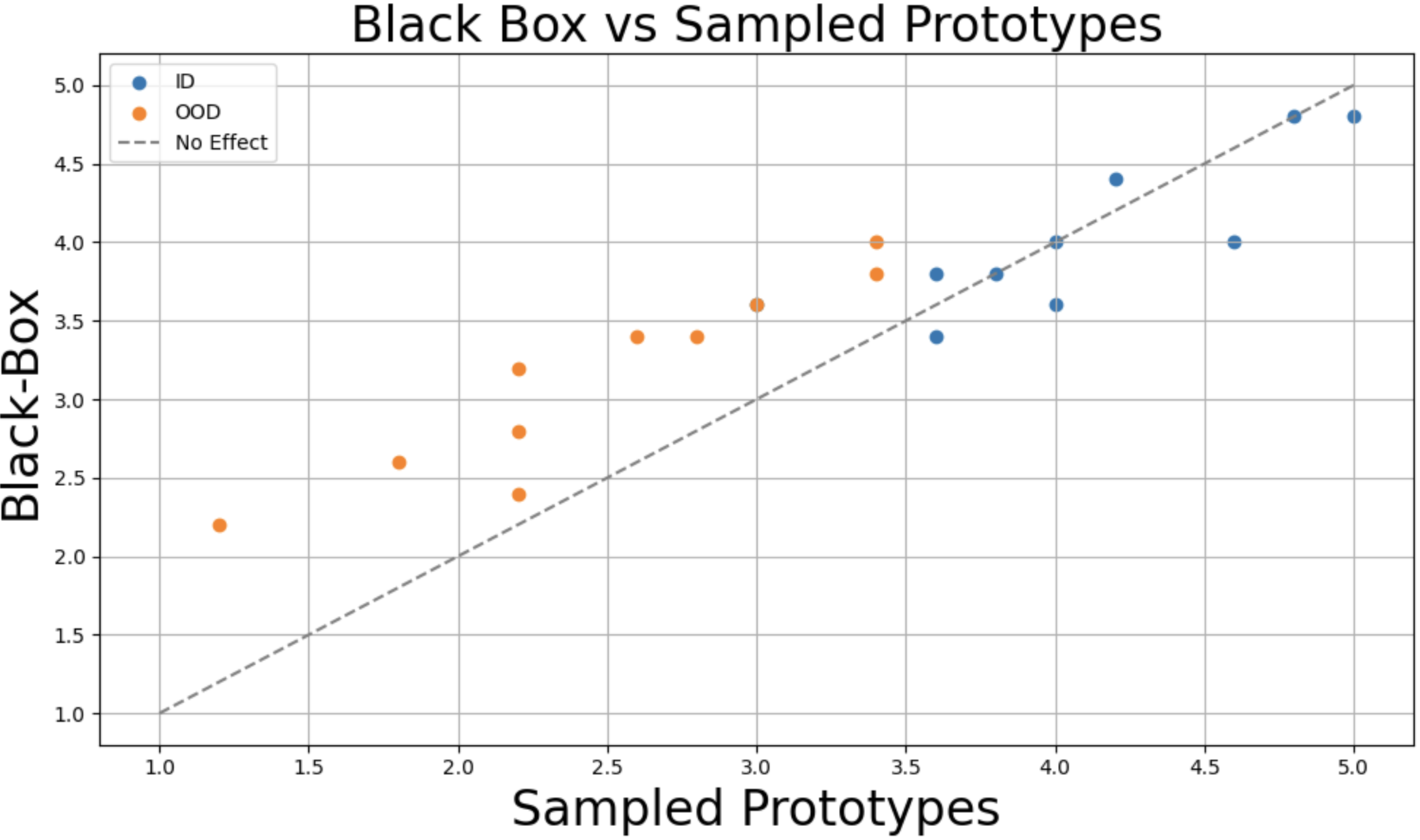}
    \caption{
    IID and OOD distribution plots for both user groups}
    \label{user_study}
\end{figure}



The interpretability of PW-Nets arises from their case-based reasoning approach, where decisions are explained through analogies to representative prototypical states. Prior work \citet{kenny2023towards} demonstrated that human-selected prototypes enable users to form accurate mental models of agent behavior, supporting effective prediction of both successes and failures. Our automated prototype selection is designed to preserve this interpretability mechanism by identifying states that capture the same decision-critical features that human experts would highlight.

To evaluate the plausibility and faithfulness of the sampled prototypes, and to analyze how prototype-based explanations influence participants’ ability to interpret and anticipate the agent’s decisions in both IID and OOD conditions, we conducted a user study in the CarRacing environment (Figure~\ref{fig:user_study}). Out of the six environments considered in our experiments, four are symbolic domains where states are represented as vectors of physical properties, while CarRacing and Atari Pong operate on raw pixels that can be visually interpreted. CarRacing was chosen because its driving actions are naturally understandable to non-expert users, making it suitable for visual inspection and evaluation \citet{rudin2021interpretablemachinelearningfundamental}.

Two groups of 25 participants were recruited. The first group interacted with PW-Nets using our sampled prototypes as global explanations, while the second group was assigned to a black-box condition in which participants were told: “The car has learned to complete the track as fast as possible in this environment by learning from millions of simulations, but no explanation is available.” In this condition, prototype images were replaced with text-only information, while the prototype group received visual exemplars that directly conveyed the agent’s reasoning process. This design isolates the contribution of prototypes to interpretability by contrasting a case-based explanation with no explanation.

Participants were presented with 20 scenarios: 10 in-distribution (ID) from the standard CarRacing-v0 environment where the agent drove safely, and 10 out-of-distribution (OOD) from a modified environment \citet{NotAnyMike_gym} introducing new road types and red obstacles that led to actual failure cases. After viewing the car’s current state and the corresponding explanatory condition, participants predicted whether the vehicle would operate safely on a five-point Likert scale. This setup assessed how well explanations enabled users to anticipate agent behavior in both familiar and novel situations.

Results are summarized in Figure~\ref{fig:user_study}. In the ID scenarios, both groups produced similar ratings, indicating that participants could reliably interpret safe behavior in either condition. In contrast, for the OOD cases where the agent failed, participants in the prototype condition were more sensitive to these failures: their ratings more closely reflected the unsafe ground truth, while the black-box group tended to overestimate safety. This demonstrates that prototype-based explanations enhance interpretability by helping users anticipate failure modes, even if they do not increase overall reported confidence.

\begin{figure}
    \centering
    \label{fig:prot_comparison}
    \includegraphics[width=1\linewidth]{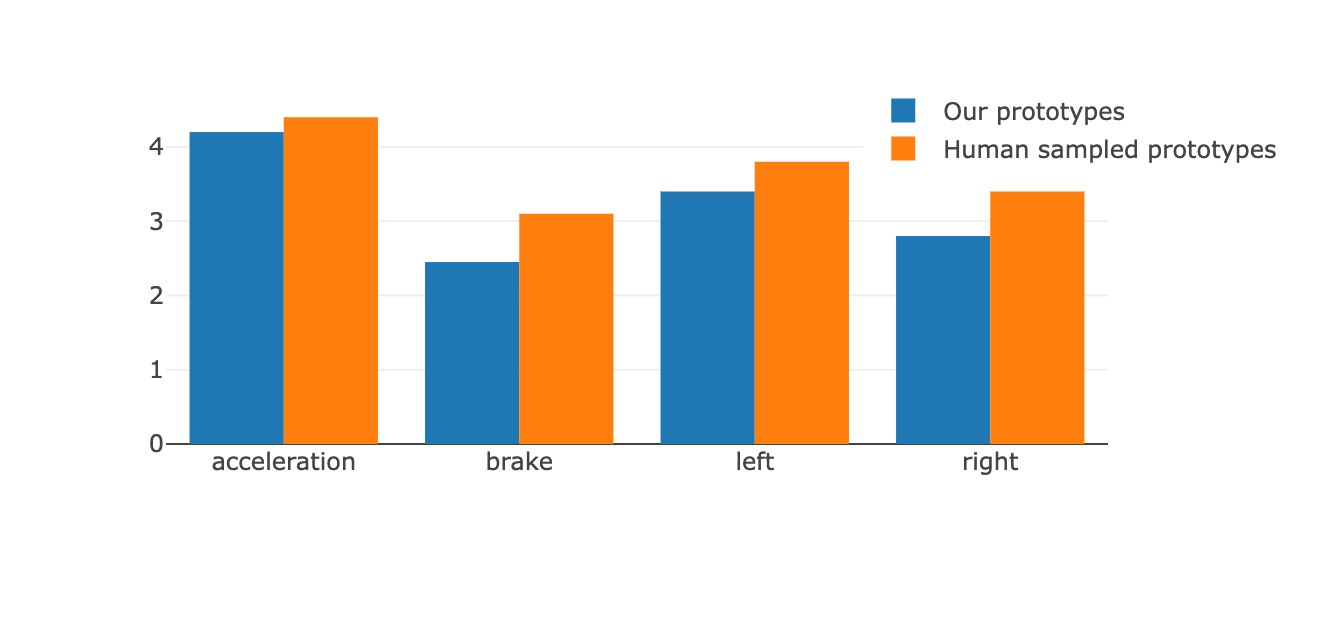}
    \caption{Comparison of Visual similarity between prototypes}
    \label{fig:placeholder}
\end{figure}

In addition, we evaluated the interpretability of our sampled prototypes relative to the human-curated prototypes used in PW-Nets (Figure~\ref{fig:prot_comparison}). For each action class, participants rated on a 1–5 scale how well the prototype represented the corresponding decision. For the acceleration class, ratings were comparable across both methods, while for the other classes human-curated prototypes were slightly preferred. While human-curated prototypes are slightly preferred in some cases, the observed differences are marginal. This suggests that our method achieves comparable interpretability in practice, while eliminating the need for manual prototype selection. Importantly, our method delivers this interpretability benefit in high-dimensional settings where human prototype selection is infeasible.

\begin{figure}[h!]
    \centering
    \includegraphics[width=\linewidth]{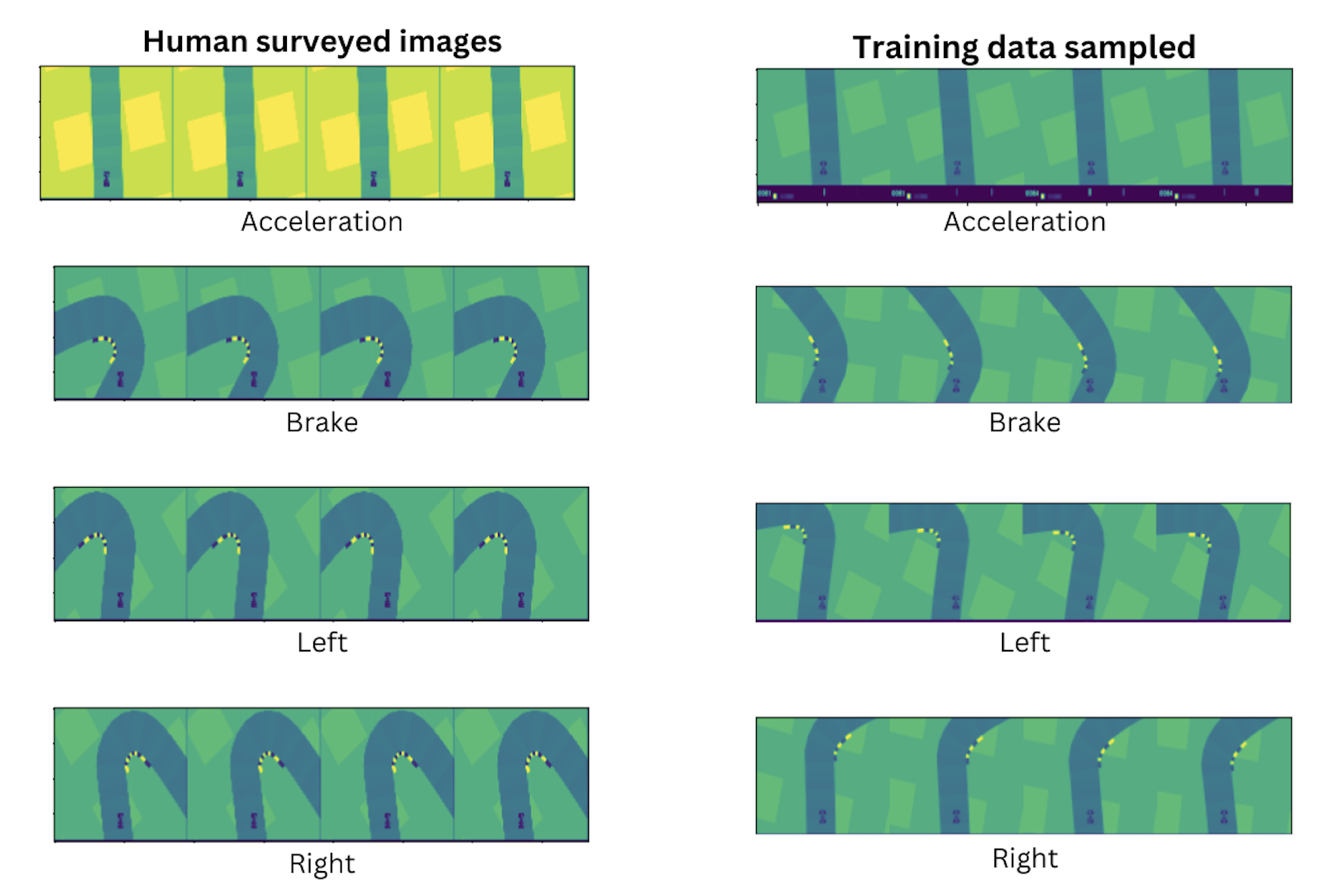}
    \caption{Visual Comparison of Human surveyed and Automated Prototypes
    }
    \label{fig: prototype_comparison}
\end{figure}

\subsubsection{\textbf{User Study details}}
Two groups of 25 users each participated in the study. All participants provided informed consent. No personally identifiable information was collected. One group was presented with the black-box model  (the "Black-Box Group") (Figure ~\ref{fig: prototype_comparison}), while the other with sampled prototypes (the "Sampled Prototypes Group") (Figure ~\ref{fig: prototype_comparison}). Both groups were given identical scenarios and instructions on how to rate them independently. The figure below shows a sample of the IID and OOD cases shown to users.

\begin{figure}[h!]
    \centering
    \includegraphics[width=\linewidth]{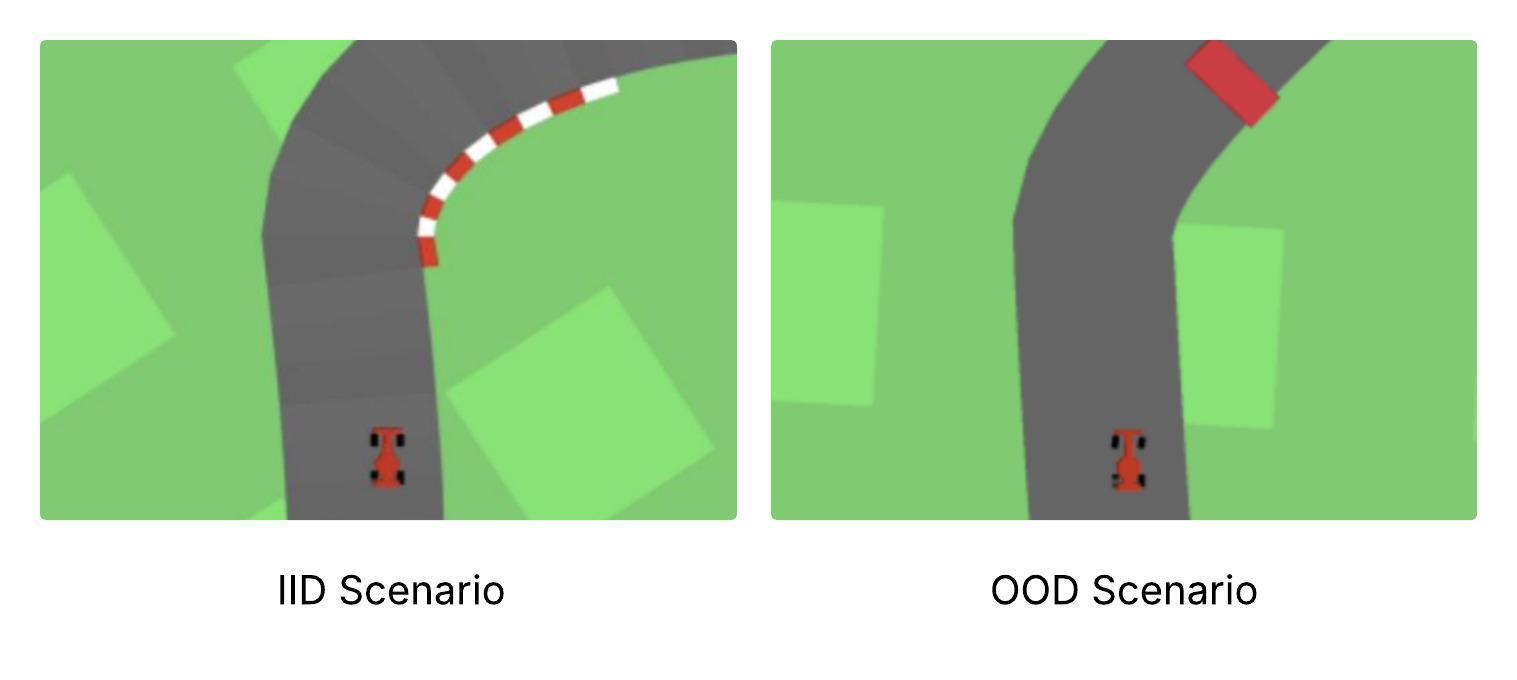}
    \caption{User Study Overview}
    \label{fig:user_study}
\end{figure}

\subsection{\textbf{Method comparisons}}
\label{sec:method_comparison}
VIPER \citet{bastani2019verifiablereinforcementlearningpolicy} extracts interpretable decision-tree policies from trained RL agents via imitation learning. Although it offers global explanations, it is limited to low-dimensional settings and does not scale well to high-dimensional or continuous control environments. The k-means method selects prototypes by choosing the cluster centers within each action class. In contrast, PWnet* learns prototypes through a joint objective that combines a clustering loss and a separation loss while simultaneously optimizing for RL performance. Moreover, our approach significantly reduces the reliance on subjective inputs, thereby promoting a more objective assessment of the prototypes. For all the environments, we used the same black-box models (Section ~\ref{sec:black_box_models}) used in PW-net.


For the Car Racing and Atari Pong environments, we recomputed the performance of the black-box models, but we retain the reported PW-Net \citet{kenny2023towards} results for CarRacing and Pong, as their evaluation depends on human-surveyed prototypes, which cannot be faithfully reproduced. For all other environments, we recompute PW-Net under identical settings. In the case of symbolic domains, we constructed canonical prototypical action-space examples, where the action of interest was set to 1 or -1 and all others to 0, and subsequently mapped these to the closest training samples. These prototypes were then used to reevaluate PW-Net’s performance across the four symbolic domains in this work.

\section{\textbf{Ablation Study}}
\label{sec:ablation_study}

To analyze the effect of each individual parameter, we have performed an ablation study on one model each from the Continuous and discrete action spaced environments. To achieve this we used the BipedalWalker and Atari pong environments respectively.

\begin{figure}[h!]
\centering
\begin{tikzpicture}

\begin{groupplot}[
    group style={
        group size=3 by 3,   
        horizontal sep=1cm,
        vertical sep=2cm,
    },
    width=0.34\textwidth,
    height=0.33\textwidth,
    grid=both,
    label style={font=\scriptsize},
    tick label style={font=\scriptsize},
    title style={font=\small},
    legend to name=combinedlegend,
]

\nextgroupplot[title={(a)$Reward$ vs $m$}, xlabel={$m$}, ylabel={$Reward$},ymin=300, ymax=320]
\addplot[thick,orange,mark=*] coordinates {
    (2,309.8) (3,310.3) (4,310.6) (5,310) (6,309) (7,306.5) (8,304)
};

\nextgroupplot[title={ (b)$Reward$ vs $\gamma$ }, xlabel={$\gamma$}, ylabel={$Reward$},ymin=300, ymax=315]
\addplot[thick,orange,mark=*] coordinates {
    (0.3,307.7) (0.5,309.4) (0.9,310.3) (0.99,310.8) (0.999,311.4)
};

\nextgroupplot[title={(c)$Reward$ vs $N_\beta$ }, xlabel={$N_\beta$}, ylabel={$Reward$},,ymin=300, ymax=315]
\addplot[thick,orange,mark=*] coordinates {
    (0.5,309) (1,309.2) (1.5,308.3) (2,307.5) (2.5,307.3) (3,306.8) 
};


\nextgroupplot[title={ (d)$Reward$ vs $N_\alpha$ }, xlabel={$N_\alpha$}, ylabel={$Reward$},ymin=300, ymax=315]
\addplot[thick,orange,mark=*] coordinates {
   (1,309.1) (2,310.24) (3,310.5) (4,310.8) (5,311) (6,311.4)
};


\nextgroupplot[title={(e)$Reward$ vs $T$ }, xlabel={$T$}, ylabel={$Reward$},ymin=300, ymax=315]
\addplot[thick,orange,mark=*] coordinates {
(0.7,308.1) (0.75,309.2) (0.8,309.3) (0.85,309.7) (0.9,310.2) (0.95,310.3)
};

\nextgroupplot[title={ (f)$Reward$ vs $\delta$ }, xlabel={$\delta$}, ylabel={$Reward$},ymin=300, ymax=315]
\addplot[thick,orange,mark=*] coordinates {
    (0.8,309.1) (1.2,309.3) (1.6,309.2) (2,309.6) (2.4,309.7) (2.8,309.5) (3.2,309.2)
};

\nextgroupplot[title={ (g)$Reward$ vs $\alpha$ }, xlabel={$\alpha$}, ylabel={$Reward$},ymin=300, ymax=315]
\addplot[thick,orange,mark=*] coordinates {
    (5,306.5) (10,303.9) (15,307.8) (20,307.3) (25,310.6) (30,311.8) (32,312.3)
};

\nextgroupplot[
    title={ (h)$Reward$ vs $\epsilon$ },
    xlabel={$\epsilon$},
    ylabel={$Reward$},
,ymin=300, ymax=315
]
\addplot[thick,orange,mark=*] coordinates {
    (0.001,310.8) (0.005,310.6) (0.05,311.1) (0.1,311.2) (0.2,311.15)
}; 


\nextgroupplot[
    title={(i)$Reward$ vs $no.of$ $prototypes$},
    xlabel={$no.of$ $prototypes$},
    ylabel={$Reward$},
    ymin=308, ymax=318,
    legend style={at={(0.5,-0.25)},anchor=north, font=\scriptsize}
]
\addplot[thick,orange,mark=*] coordinates {
    (1,310.8) (5,312.3) (10,312.9) (15,312.1) (20,313.4) (25,313.8) (30, 313.95) (35,313.9) (40,313.85)
};

\end{groupplot}
\end{tikzpicture}

\caption{Ablation study on BipedalWalker environment}
\label{fig:abilation_biped}
\end{figure}
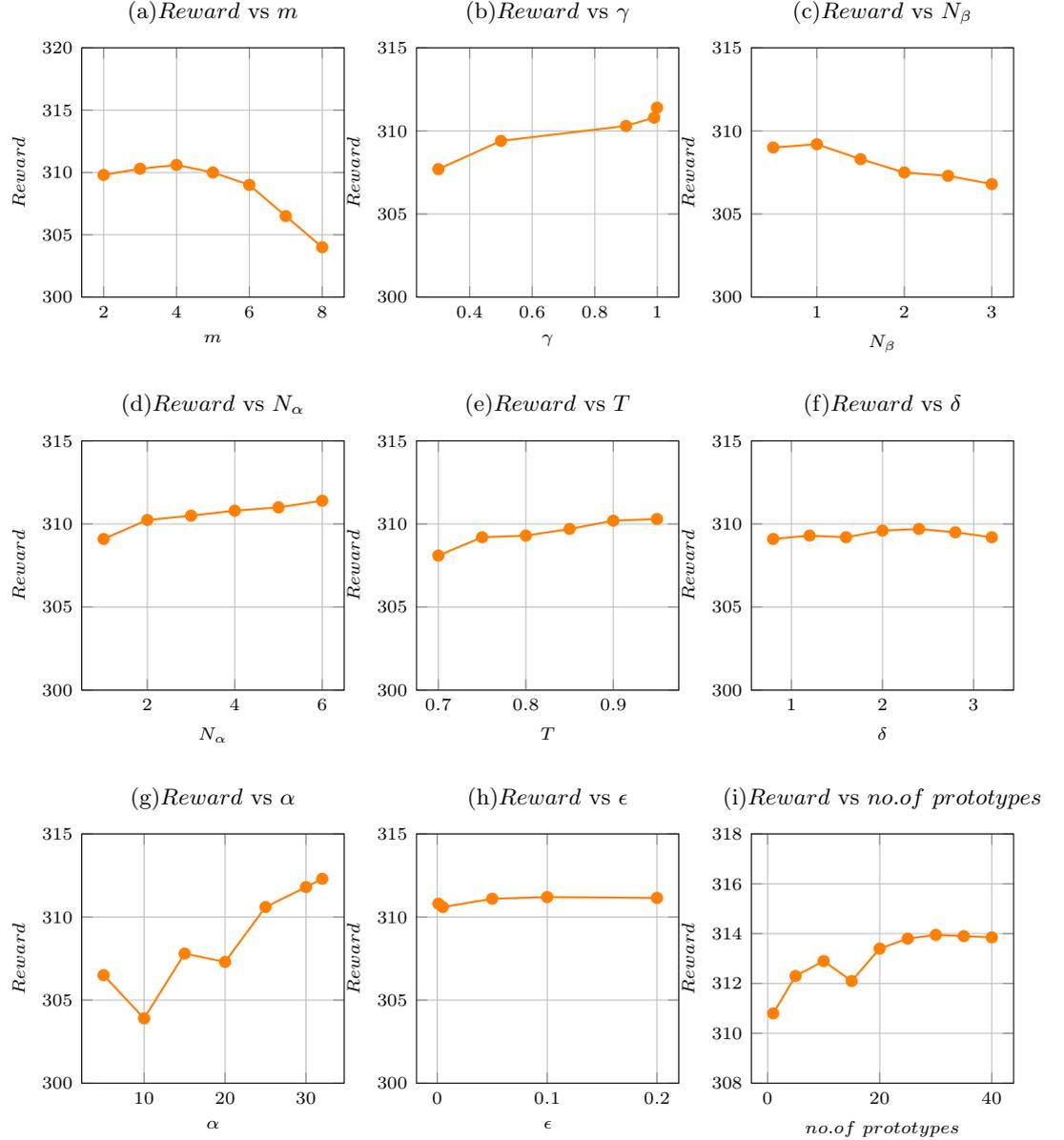

\begin{figure}[h!]
\centering
\begin{tikzpicture}

\begin{groupplot}[
    group style={
        group size=3 by 3,   
        horizontal sep=1cm,
        vertical sep=2cm,
    },
    width=0.34\textwidth,
    height=0.33\textwidth,
    grid=both,
    label style={font=\scriptsize},
    tick label style={font=\scriptsize},
    title style={font=\small},
    legend to name=combinedlegend,
]

\nextgroupplot[title={(a)$Reward$ vs $m$}, xlabel={$m$}, ylabel={$Reward$},ymin=5, ymax=20]
\addplot[thick,blue,mark=*] coordinates {
    (2,13.8) (3,13.9) (4,13.3) (5,12.6) (6,12.5) (7,11.9) (8,11.5)
};

\nextgroupplot[title={ (b)$Reward$ vs $\gamma$ }, xlabel={$\gamma$}, ylabel={$Reward$},ymin=5, ymax=20]
\addplot[thick,blue,mark=*] coordinates {
    (0.3,12.87) (0.5,13.1) (0.9,13.6) (0.99,14.29) (0.999,14.38)
};

\nextgroupplot[title={(c)$Reward$ vs $N_\beta$ }, xlabel={$N_\beta$}, ylabel={$Reward$},ymin=10, ymax=16]
\addplot[thick,blue,mark=*] coordinates {
    (0.5,13.4) (1,13.3) (1.5,13.2) (2,12.9) (2.5,12.7) (3,12.6) 
};


\nextgroupplot[title={ (d)$Reward$ vs $N_\alpha$ }, xlabel={$N_\alpha$}, ylabel={$Reward$},ymin=10, ymax=16]
\addplot[thick,blue,mark=*] coordinates {
   (1,12.6) (2,12.9) (3,13.1) (4,13.4) (5,13.6) (6,13.85)
};


\nextgroupplot[title={(e)$Reward$ vs $T$ }, xlabel={$T$}, ylabel={$Reward$},ymin=10, ymax=16]
\addplot[thick,blue,mark=*] coordinates {
(0.7,12.5) (0.75,12.7) (0.8,12.85) (0.85,12.91) (0.9,13.05) (0.95,13.06)
};

\nextgroupplot[title={ (f)$Reward$ vs $\delta$ }, xlabel={$\delta$}, ylabel={$Reward$},ymin=10, ymax=16]
\addplot[thick,blue,mark=*] coordinates {
    (0.8,12.2) (1.2,12.31) (1.6,12.2) (2,12.4) (2.4,12.45) (2.8,12.5) (3.2,12.35)
};

\nextgroupplot[title={ (g)$Reward$ vs $\alpha$ }, xlabel={$\alpha$}, ylabel={$Reward$},ymin=10, ymax=16]
\addplot[thick,blue,mark=*] coordinates {
    (5,11.8) (10,11.5) (15,13.56) (20,12.9) (25,13.9) (30,14.8) (32,14.85)
};

\nextgroupplot[
    title={ (h)$Reward$ vs $\epsilon$ },
    xlabel={$\epsilon$},
    ylabel={$Reward$},
,ymin=10, ymax=16
]
\addplot[thick,blue,mark=*] coordinates {
    (0.001,11.8) (0.005,11.6) (0.05,11.36) (0.1,11.45) (0.2,11.49)
}; 

\nextgroupplot[
    title={(i)$Reward$ vs $no.of$ $prototypes$},
    xlabel={$no.of$ $prototypes$},
    ylabel={$Reward$},
    ymin=10, ymax=16,
    legend style={at={(0.5,-0.25)},anchor=north, font=\scriptsize}
]
\addplot[thick,blue,mark=*] coordinates {
   (1,13.36) (2,13.57) (4,13.94) (6,14.39) (8,14.76) (10,15.34) (12,15.1) (14,13.7) (16,13.2) (18,12.8) (20,12.37)
};

\end{groupplot}
\end{tikzpicture}

\caption{Ablation study on Atari Pong environment}
\label{fig:abilation_atari}
\end{figure}
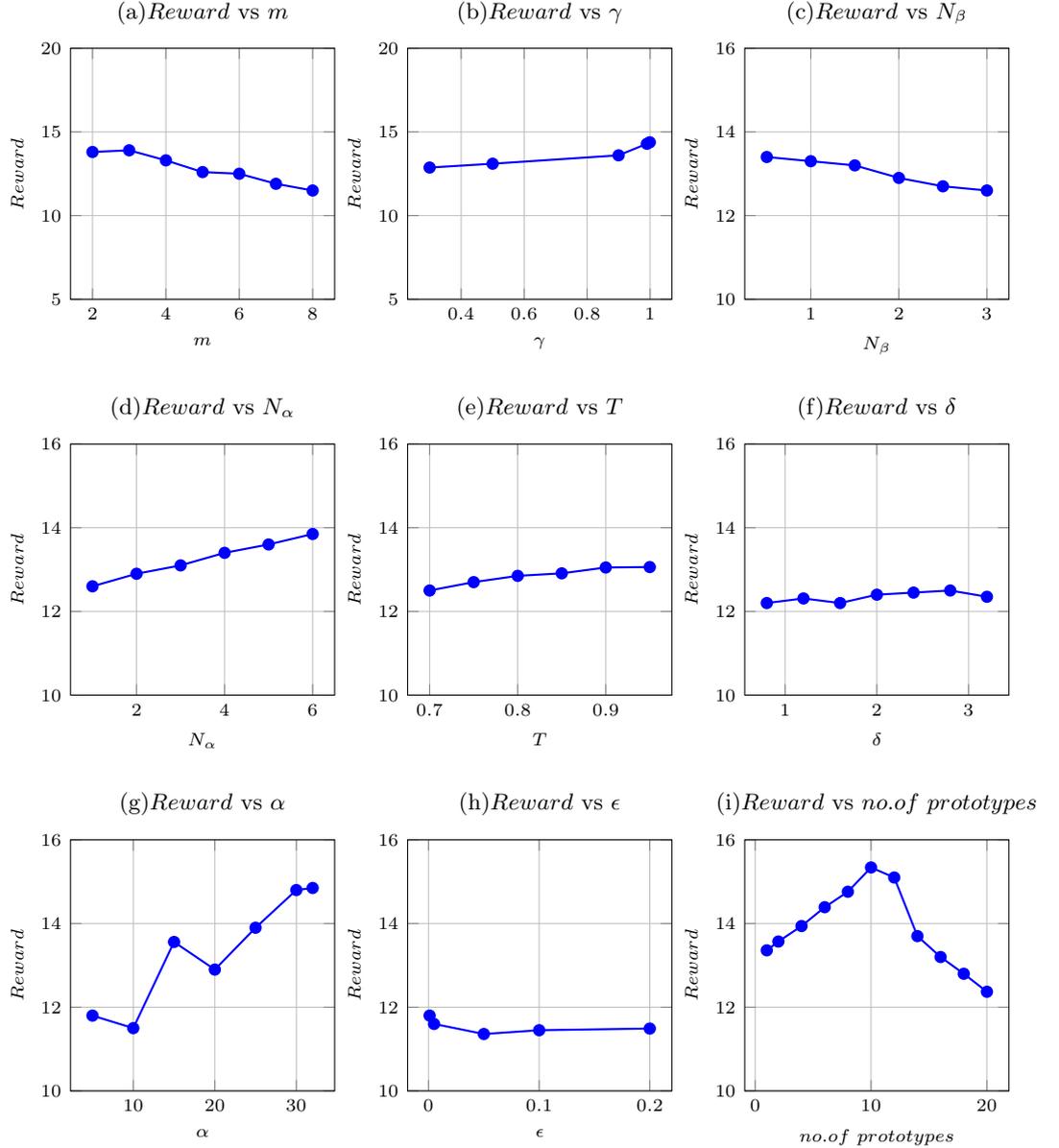

\subsection{\textbf{Effect of \texorpdfstring{$m$}{m}}}
The parameter $m$ denotes the dimension of the linear submanifold $X_i$, which locally approximates the data manifold around a point $h_\theta(z)$. 
To examine its effect, we vary $m$ in the range $[2,8]$ with a step size of $1$. 
As shown in (Figure~\ref{fig:abilation_atari} and  Figure~\ref{fig:abilation_biped})(a), performance consistently decreases in both the environments as $m$ increases. 
This trend arises because $X_i$ is intended to approximate the immediate neighborhood of a point, which is inherently low-dimensional. 
Larger values of $m$ may lead to overfitting, since only a limited number of nearby samples are available within a batch to reliably estimate $X_i$, thereby degrading performance. 
Furthermore, we observe that the computational overhead for prototype sampling increases with larger $m$, underscoring the trade-off between accuracy and efficiency.

\subsection{\textbf{Effect of \texorpdfstring{\(\gamma\)}{gamma}}}
The parameter $\gamma$ denotes the momentum constant used to update the proxy vector $\theta_m$ during prototype sampling. 
Following \citet{he2020momentumcontrastunsupervisedvisual}, higher values of $\gamma$ are expected to yield improved performance, as the proxy updates become smoother and more stable. 
Consistent with this observation, (Figure~\ref{fig:abilation_atari} and Figure~\ref{fig:abilation_biped})(b) shows that in both models, performance improves as $\gamma$ increases, highlighting the importance of stable momentum updates for effective representation learning.

\subsection{\textbf{Effect of \texorpdfstring{$N_\alpha$ \& $N_\beta$}{Nalpha \& Nbeta}}} 
The parameters $N_\alpha$ and $N_\beta$ control the decay of similarity based on the orthogonal and projected distances, respectively, of a point from the linear submanifold in the neighborhood of another point. 
We vary $N_\alpha$ in the range $[1,6]$ with a step size of $1$, and $N_\beta$ in the range $[0.5,3]$ with a step size of $0.5$. 
As shown in (Figure ~\ref{fig:abilation_atari} and Figure~\ref{fig:abilation_biped})(c), increasing $N_\beta$ leads to decrease in performance in both the environments.
In contrast, (Figure~\ref{fig:abilation_atari} and Figure ~\ref{fig:abilation_biped})(d) shows that performance improves with larger $N_\alpha$ in both the environments.

This effect can be explained by the relationship between $N_\alpha$ and $N_\beta$: as $N_\alpha$ approaches $N_\beta$, a point $A$ at distance $\varepsilon$ within the linear neighborhood of a point $B$ (and thus sharing many features with $B$ and its neighbors) may be treated as equally dissimilar to $B$ as another point $C$ located at an orthogonal distance $\varepsilon$ from the neighborhood of $B$. In the experiments when $N_\beta$ was varied $N_\alpha$ is set to 4, as $N_\beta$ increases from 0.5 to 3 it becomes closer to $N_\alpha$ which is leading to a decrease in performance. When $N_\alpha$ was varied from 1 to 6 $N_\beta$ was set to 0.5, as $N_\alpha$ increases from it becomes larger than $N_\beta$ which is leading to an increase in performance.

\subsection{\textbf{effect of \texorpdfstring{$T$}{T}}}
The reconstruction threshold $T$ determines the quality of points admitted into the linear submanifold $X_i$. 
We vary $T$ in the range $[0.7,0.95]$ with a step size of $0.05$. 
As shown in (Figure~\ref{fig:abilation_atari} and Figure~\ref{fig:abilation_biped})(e), the models in both environments exhibit a clear upward trend in performance as $T$ increases, underscoring the importance of ensuring that only high-quality points are incorporated into $X_i$.

\subsection{\textbf{Effect of \texorpdfstring{$\delta$}{delta}}}
The scaling factor $\delta$ regulates the maximum separation between dissimilar points. We vary $\delta$ in the range $[0.8, 3.2]$ with a step size of $0.4$. As shown in (Figure ~\ref{fig:abilation_atari} and Figure~\ref{fig:abilation_biped})(f), the performance remains relatively stable across this range in both environments, highlighting the robustness of our method.

\subsection{\textbf{Effect of \texorpdfstring{$\alpha$}{alpha }}}
The scaling factor $\alpha$ controls the sharpness of the exponential term in the Proxy Anchor loss. We vary its value over ${5, 10, 15, 20, 25, 30, 32}$. As shown in (Figure ~\ref{fig:abilation_atari} and Figure~\ref{fig:abilation_biped})(g), models in both environments exhibit an overall increasing trend in performance with larger $\alpha$.

\subsection{\textbf{Effect of \texorpdfstring{$\epsilon$ }{epsilon}}}

The margin parameter $\epsilon$ enforces that positive embeddings are pulled within this distance from their corresponding class proxies. We vary its value across ${0.001, 0.005, 0.05, 0.1, 0.2}$. As shown in (Figure~\ref{fig:abilation_atari} and Figure~\ref{fig:abilation_biped})(h), models in both the environments demonstrate stable performance across the range of $\epsilon$, undermining its effect in the loss function.

\subsection{\textbf{Effect of \texorpdfstring{$no.of$ $prototypes$ }{no.of prototypes}}}

To investigate the effect of prototype count on performance, we conducted an ablation study in the Bipedal Walker and Atari Pong environments. In Bipedal Walker ~\ref{fig:abilation_biped}(i), rewards consistently increased with additional prototypes until reaching a plateau. In contrast, in Atari Pong ~\ref{fig:abilation_atari}(i), rewards initially improved with more prototypes but began to decline beyond a certain point. We attribute this divergence to differences in state representation.

Bipedal Walker is a symbolic domain where states encode physical properties such as position and velocity, providing relatively low-noise inputs. By comparison, Atari Pong represents states as raw pixels, which must be encoded by a neural network before prototype selection. This pixel-based encoding introduces noise, and as the number of prototypes increases, the accumulated noise degrades performance.

\section{\textbf{Conclusion and Future work}}
The application of Deep Reinforcement Learning (Deep RL) spans from automated 
game simulations to fine-tuning large language models (LLMs) using preference 
data. However, in the absence of transparency regarding the agent's actions and 
intentions, deploying such systems in high-stakes or sensitive domains remains 
impractical \citet{rudin2019stopexplainingblackbox}. PW-Net addresses this 
challenge by providing interpretability for deep RL agents through example-based 
reasoning using human-understandable concepts. While relying on human-annotated 
prototypes offers valuable insights, it is not feasible across all domains. To 
overcome this limitation, our approach automatically samples prototypes from the 
training data itself, leveraging the geometric structure of the encoded state 
space to select representative and discriminative exemplars without requiring 
expert curation. Through user studies, we demonstrate that trust in the model's 
behavior especially under out-of-distribution (OOD) scenarios where failures 
are likely can be effectively assessed using our automatically sampled 
prototypes, with participants better anticipating agent failures compared to 
a black-box baseline.

Several promising directions remain open for future work. First, while our 
method supports a variable number of prototypes per class with performance 
scaling behavior analyzed across symbolic and pixel-based domains in 
Section~\ref{sec:ablation_study} extending the prototype discovery framework to 
output spaces of significantly larger cardinality, such as those encountered 
in generative language modeling, presents an important challenge. In such 
settings, the action space may grow to vocabulary scale, and efficient 
prototype selection under these conditions would require hierarchical or 
clustering-based strategies that preserve the geometry-aware properties of 
our current approach. \citet{xie2023protolmprototypicalnetworkbasedframework} 
made initial progress in this direction for sentence classification, but 
prototype-based interpretability for open-ended generation remains largely 
unsolved. Second, our current framework operates post-hoc on a fixed 
pre-trained policy; an interesting extension would be to investigate whether 
geometry-aware prototype discovery can be incorporated as a soft regularizer 
during policy training itself, potentially guiding the encoder to produce 
representations that are both task-optimal and inherently more amenable to 
prototype-based explanation. Third, the piecewise-linear manifold construction 
currently operates on encoded state representations from a single policy. 
Extending this to multi-task or transfer learning settings where a shared 
encoder serves multiple policies across environments could yield prototypes 
that capture transferable, task-agnostic behavioral primitives, broadening the 
scope of interpretability beyond individual environment instances. Finally, 
evaluating our approach in higher-stakes real-world domains such as autonomous 
driving simulation and robotic manipulation, where prototype-based explanations 
could directly support human oversight and intervention, represents a natural 
and impactful next step.

\section{Acknowledgments}

The authors were partially supported by the US National Science Foundation under awards IIS-2520978, GEO/RISE-5239902, the Office of Naval Research Award N00014- 23-1-2007, DOE (ASCR) Award DE-SC0026052, and the DARPA D24AP00325-00. Approved for public release; distribution is unlimited.

\bibliographystyle{plainnat}
\bibliography{iclr2026_conference}

\appendix

\end{document}